\documentclass[a4paper]{article}

\usepackage{INTERSPEECH2021}
\usepackage{amsmath,graphicx}

\title{Super-Human Performance in Online Low-latency Recognition of Conversational Speech}

\name{Thai-Son Nguyen$^{12}$, Sebastian St\"uker$^{12}$, Alex Waibel$^{12}$}

\address{Institute for Anthropomatics and Robotics, 
Karlsruhe Institute of Technology}
\address{$^1$ Institute for Anthropomatics and Robotics, Karlsruhe Institute of Technology \\
$^2$ Karlsruhe Information Technology Solutions --- kites GmbH}
\email{firstname.lastname@kit.edu}
\begin{document}

\maketitle
\begin{abstract}
Achieving super-human performance in recognizing human speech has been a goal for several decades as researchers have worked on increasingly challenging tasks. In the 1990's it was discovered, that conversational speech between two humans turns out to be considerably more difficult than read speech as hesitations, disfluencies, false starts and sloppy articulation complicate acoustic processing and require robust joint handling of acoustic, lexical and language context. Early attempts with statistical models could only reach word error rates (WER) of over 50\% which is far from human performance with shows a WER of around 5.5\%. Neural hybrid models and recent attention-based encoder-decoder models have considerably improved performance as such contexts can now be learned in an integral fashion. However, processing such contexts requires an entire utterance presentation and thus introduces unwanted delays before a recognition result can be output.  In this paper, we address performance \emph{as well as} latency.  We present results for a system that can achieve super-human performance, i.e. a WER of 5.0\% on the Switchboard conversational benchmark, at a word based latency of only 1 second behind a speaker's speech. The system uses multiple attention-based encoder-decoder networks integrated within a novel low latency incremental inference approach.
\end{abstract}

\vspace{0.5cm}
\section{Introduction}
\label{sec:intro}
Sequence-to-sequence (S2S) attention-based models \cite{chorowski2015attention,chan2016listen} are a currently one of the best performing approaches to end-to-end automatic speech recognition (ASR). A lot of research has already been dedicated to boost the performance of S2S models.
Several works \cite{chiu2018state,zeyer2018improved,weng2018improving,pham2019very,nguyen2020improving} have successfully pushed up the state-of-the-art performance records on different speech recognition benchmarks and proved the superior performance of S2S models over conventional speech recognition models in an offline setting. As so, the next research trend is to apply S2S speech recognition in practice. Many practical applications need to work ASR systems in real-time run-on mode with low-latency \cite{nguyen2020low,niehues2016dynamic}. 

Early studies \cite{jaitly2016online,raffel2017online,chiu2017monotonic} pointed out that the disadvantage of an S2S model used in online condition lies in its attention mechanism, which must perform a pass over the entire input sequence for every output element. \cite{raffel2017online,chiu2017monotonic} have dealt with this disadvantage by proposing a so-called monotonic attention mechanism that enforces a monotonic alignment between the input and output sequence. Later on, \cite{fan2019online,miao2019online,tsunoo2019towards} have additionally resolved the latency issue of bidirectional encoders by using efficient chunk-based architectures. More recent works \cite{miao2020transformer,moritz2020streaming,nguyen2020high,wu2020streaming,zhang2020streaming,kumar20201} have addressed these latency issues for different S2S architectures.

While most of the studies focus on model modifications to make S2S models capable of online processing with minimal accuracy reduction, they lack thoughtful research on the latency aspect. In this work, we analyze the latency that the users suffer while interacting with an online speech recognition system, and propose to measure it with two separate terms \textit{computation latency} and \textit{confidence latency}. While computation latency reflects the common real-time factor (RTF), confidence latency corresponds to the time an online recognizer needs to confidently decide its output. We show that with the support of new computing hardware (such as GPUs), the computation latency of S2S models is relatively small (even for big models), and the confidence latency is a more critical criterion which, for the first time, we address thoroughly.

To optimize for confidence latency, we consider the online processing of S2S models as an incremental speech recognition problem. We propose an \emph{incremental inference} approach with two stability detection methods to convert an S2S model to be used in online speech recognition and to allow to trade-off between latency and accuracy. Our experimental results show that it is possible to use a popular Long Short-Term Memory (LSTM)~\cite{hochreiter1997long} or self-attention based S2S ASR model for run-on recognition without any model modification. With a delay of 1.8 seconds in all output elements, all the experimental models retain their state-of-the-art performance when performing offline inference. Our best online system, which successfully employs three S2S models in a low-latency manner, achieves a word-error-rate (WER) of 5.0\% on the Switchboard benchmark. To the best of our knowledge, this online accuracy is at par with the state-of-the-art offline performance. We also demonstrate that it is possible to achieve human performance as measured in \cite{xiong2016achieving,saon2017english} while producing output at very low latency.

\begin{figure*}[th]
	\centering
	\includegraphics[width=0.72\linewidth]{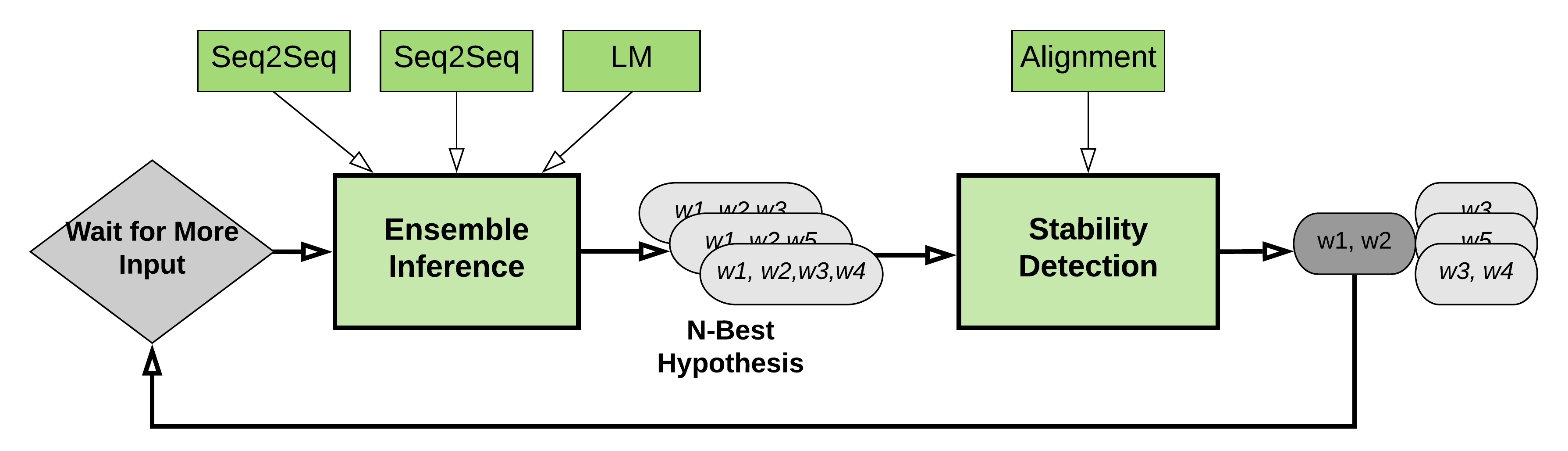}
	\vspace{-0.1cm}
	\caption{Incremental inference for low-latency S2S ASR}
	\label{fig:architecture}
    \vspace{-0.0cm}
\end{figure*}

\vspace{0.4cm}
\section{Sequence-to-sequence Based Low-latency ASR}
\label{sec:framework}
In this section, we first describe different sequence-to-sequence ASR architectures investigated in this paper. We then present the proposed incremental inference with two stability detection methods.

\vspace{-0.0cm}
\subsection{Models}
There have been two efficient approaches for making S2S ASR systems. The first approach employs LSTM layers in both encoder and decoder networks, while the second follows the Transformer architecture \cite{vaswani2017attention} which uses solely self-attention modules to construct the whole S2S network. In this work, we investigate both of the S2S architectures for the online low-latency setting.

Our LSTM-based S2S model employs two time-delay neural network (TDNN) layers \cite{waibel1989tdnn,lecun2015deep} with a total time stride of four for down-sampling followed by several bidirectional LSTM layers to encode the input spectrogram. In the decoder, we adopt two layers of unidirectional LSTMs for modeling the sequence of sub-word labels and the multi-head \textit{soft-attention} function proposed in \cite{vaswani2017attention} to generate attentional context vectors. In detail, the LSTM-based model works as the following neural network functions:
\begin{eqnarray*}
	enc = LSTM(TDNN(spectrogram)) \\
	emb = LSTM(Embedding(symbols)) \\
	ctx = SoftAttention(emb, enc, enc) \\
	y = Softmax(ctx + emb)
\end{eqnarray*}

In the Transformer model, the down-sampling is handled by a linear projection layer on four consecutive stacked feature vectors. The rest of the model architecture is similar to the Transformer approach proposed for speech recognition in \cite{pham2019very}. We also adopt the layer stochastic technique to efficiently employ more self-attention layers in both encoder and decoder.

For more details of the model architectures and offline evaluations, we would refer the readers to \cite{nguyen2020improving} and \cite{pham2019very}.

\vspace{0.2cm}
\subsection{Incremental Inference}
\label{ssec:incl_infer}
Figure \ref{fig:architecture} illustrates our proposed architecture that allows S2S models to produce incremental transcriptions on a speech stream. In the architecture, we handle the two tasks of inference and stability detection by two separate components in a processing pipeline. The first step in the pipeline is to wait for a chunk of acoustic frames with a predefined length (i.e., 200ms), which is then sent to the inference component. The inference component needs to accumulate all the chunks received so far and extend the current \textit{stable} hypothesis to produce a set of new \textit{unstable} hypotheses. This unstable set is then provided to the \textit{stability detection} component for detecting a longer stable hypothesis.

As the stability detection is handled separately, we are able to involve multiple models for the inference to improve recognition accuracy. The involved models can be S2S models with different architectures or language models trained on different text data. All of these models can be uniformly combined via the ensemble technique.

\vspace{0.2cm}
\subsection{Stability Detection}
\label{ssec:stability_dectect}
Stability detection is the key to make the system work in the incremental manner and to produce low latency output. For an HMM based speech recognition system, stability conditions can be determined incrementally during the time-synchronous Viterbi search \cite{brown1982partial,wachsmuth1998integration,selfridge2011stability}. Due to lack of time alignment information and unstable internal hidden states (e.g., of a bidirectional encoder), it is not straightforward to apply the same idea to S2S models. In this work, we investigate a combination of two stability detection conditions for incremental S2S speech recognition:

\begin{itemize}
    \item \textbf{Shared prefix in all hypotheses}: Similar to the \textit{immortal prefix} \cite{brown1982partial,selfridge2011stability} in HMM ASR, this condition happens when all the active hypotheses in the beam-search share the same prefix. However, different from HMM ASR, this condition may not strongly lead to an \textit{immortal} partial hypothesis due to the unstable search network states in S2S beam-search.
    \item \textbf{Best-ranked prefix with reliable endpoint}: Since it may require a long delay for a \textit{shared prefix} to happen, we also consider a different approach to improve the latency. We make use of the observation from \cite{wachsmuth1998integration} for HMM ASR, that the longer a prefix remains to be part of the most likely hypothesis, the more stable it is. Applied to S2S models, we need a method to align a prefix with audio frames, and so be able to find its endpoint in time. We follow the approach in \cite{nguyen2020high} for the estimation of a prefix endpoint. First, this approach requires to train a single-head attention LSTM-based S2S model with the attention-based constraint loss \cite{nguyen2020high}. Then, the endpoint of a prefix $C$ is estimated during incremental inference by finding a time $t_c$ such that the sum of all attention scores from the covering window $[0, t_c]$ is at least $0.95$. After that, we can measure $\Delta$ as the difference between the estimated endpoint and the end of the audio stream. $\Delta$ will be used as the single input to decide the prefix $C$ is reliable enough and considered to output.
\end{itemize}

\vspace{-0.0cm}
\section{Measure of Latency}
\label{sec:latency_measure}
Latency is one of the most important factors that decide the usability of an user-based online ASR system. A latency measure needs to reflect the actual delay that the users perceive so that the improvement of latency can lead to better usability. Strictly, the latency observed by a user for a single word is the time difference between when the word was uttered and when its transcript appeared to the user and will never be changed again. We formulate this complete latency as follows.

Let's assume a word $w$ has been uttered, i.e., completely pronounced, at time $U_w$. Let $C_w$ be the time that the ASR system can start to process the audio of $w$ and that the ASR system can \textit{confidently} infer $w$ after a \textit{delay} of $D_{w}$, the time needed to perform the inference. The user-perceived latency with regard to $w$ is then:
\begin{eqnarray*}
	Latency_w = C_{w} + D_{w} + T_{w} - U_{w}
\end{eqnarray*}
\noindent where $T_{w}$ presents the transmitting time for audio and text data. $T_{w}$ is usually small and can be omitted.

For a speech utterance $S$ consisting of $N$ words $w_1$, $w_2$,.. $w_N$, we are interested in the average latency:
\begin{eqnarray*}
	Latency_S = \sum \limits_{i}^{N} (D_{w_i} + C_{w_i} - U_{w_i}) / N \\
	          = \sum \limits_{i}^{N} D_{w_i} / N + \sum \limits_{i}^{N} C_{w_i} / N - \sum \limits_{i}^{N} U_{w_i} / N \\
	          = \sum \limits_{i}^{N} D_{w_i} / N + \sum \limits_{i}^{N} C_{w_i} / N - \sum \limits_{i}^{N} (U_{w_i}-\delta) / N + \delta \\
	          = D_{avg} + C_{avg} - U_{avg-\delta} + \delta
\end{eqnarray*}
In the final equation, the first term represents the computational delay. If we normalize this term by length of the decoding audio segments, then we obtain the real-time factor of the ASR system. The second term indicates how much acoustic evidence the model needs to confidently decide on its output. This latency term makes the difference in calculating the latency for online vs. offline processing. For offline processing, it is always a constant for a specific test set, since all the offline transcripts are output at the end of the test set.

To estimate the third term, we usually need to use an external time alignment system, e.g. a Viterbi alignment using an Hidden Markov Model (HMM) based acoustic model. It is inconvenient to re-run the time alignment for every new transcript. To cope with this issue, \cite{nguyen2020high} introduced a fixed delay $\delta$ for all the outputs, and proposed to pre-compute a set of $U_{avg-\delta}$ for different $\delta$. Later on, only the calculation of $C_{avg}$ is required as the average delay can be found by comparing $C_{avg}$ with the pre-computed set.

The latency improvement requires the optimization of both terms $D_{avg}$ and $C_{avg}$ which we refer to as \textit{computation latency} and \textit{confidence latency}. While computation latency can be improved by faster hardware or improved implementations for the search, confidence latency mainly depends on the recognition model.

\vspace{-0.0cm}
\section{Experimental Setup}
\label{sec:setups}
Our experiments were conducted on the Fisher+Switchboard corpus consisting of 2,000 hours of telephone conversation speech. The Hub5'00 evaluation data was used as the test set, reporting separate performance numbers for the Switchboard (SWB) and CallHome (CH) portions.

All our models use the same input features of 40 dimensional log-mel filterbanks to predict 4,000 byte-pair-encoded (BPE) sub-word units. During training, we employ the combination of two data augmentation methods \textit{Dynamic Time Stretching} and \textit{SpecAugment} \cite{nguyen2020improving} to reduce model overfitting. Adam with an adaptive learning rate schedule is used to perform 200,000 updates. The model parameters of the 5 best epochs according to the perplexity on the cross-validation set are averaged to produce the final model.

\begin{figure}[t]
	\centering
	\vspace{-0.0cm}
	\includegraphics[width=0.85\linewidth]{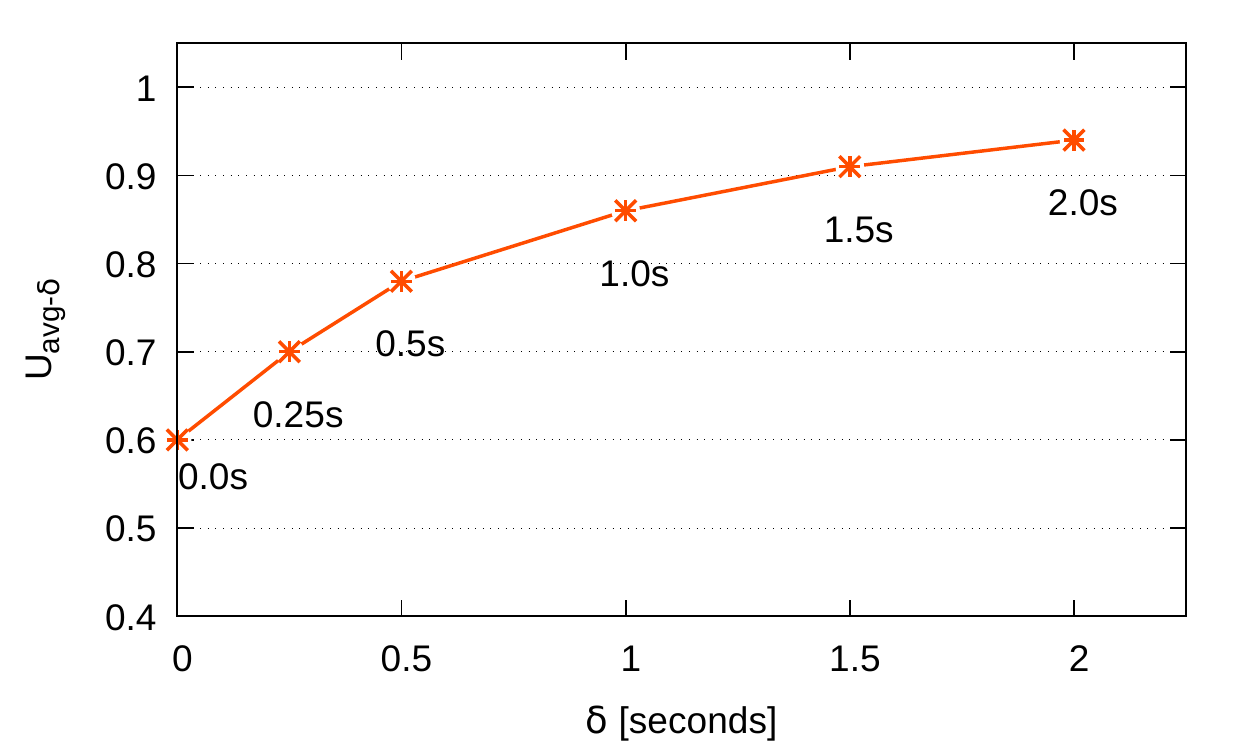}
    \vspace{-0.0cm}
	\caption{Confidence latency conversion.}
	\label{fig:confidence_conversion}
	\vspace{-0.2cm}
\end{figure}

\vspace{-0.0cm}
\subsection{Latency Evaluation}
\label{ssec:latency_eval}
We evaluate our systems with the decomposed latency terms from Section \ref{sec:latency_measure}. \textit{Computation latency} is measured every time when incremental inference is performed, while for \textit{confidence latency} we adopt a similar approach to \cite{nguyen2020high} to estimate the terms $C_{avg}$ and $U_{avg}$. First, we build a good HMM-based force-alignment system and use it to find time alignment for the test set transcripts. $U_{avg}$ is calculated as the average of the ending times of all the transcript words found by the alignment system. To normalize $U_{avg}$ between 0 and 1, all the time alignment indexes are divided by their utterance lengths.
We then shift the time indexes to the right with different $\delta$ (the delay term) to compute $U_{avg-\delta}$. This results in a conversion chart illustrated in Figure \ref{fig:confidence_conversion}. Later on, $C_{avg}$ is computed the same way for the systems, and the corresponding delay is extracted from the conversion chart.

\vspace{0.2cm}
\begin{table}[h]
	\caption{Experimental systems and their offline accuracy. The optimal beam size of 8 was found for all the systems.}
	\label{tab:systems}
	\setlength{\tabcolsep}{4pt}
	\centering
	\begin{tabular}{|c|c|c|c|c|}
		\hline
		ID & Model Type & \#Params & SWB & CH \\
		\hline \hline
		\textit{S1} & 6x2 LSTM-1024 & 162M & 5.8 & 11.8 \\
		S2 & 6x2 LSTM-1536 & 258M & 5.3 & 11.5 \\
		T1 & 24x8 Transformer & 111M & 5.8 & 11.9 \\
		\hline
		E1 & S1 + S2 & 420M & 5.3 & 10.9 \\
		E2 & S1 + S2 + T1 & 531M & 5.0 & 10.1 \\
		\hline
	\end{tabular}
\end{table}
\vspace{-0.0cm}

\vspace{0.0cm}
\section{Results}
\label{sec:results}
\vspace{-0.0cm}
\subsection{Models and Offline Accuracy}
\label{ssec:result_systems}
We constructed two LSTM-based models with different model sizes. The smaller one uses 1-head attention and was trained with the attention-based constraint loss proposed in \cite{nguyen2020high} to prevent the attention function from using future context, while the bigger uses 8-head attention and produces better accuracy. The smaller model \textit{S1} can be used either for inference or to extract the endpoint of a hypothesis prefix following \cite{nguyen2020high}. Additionally, we experiment with a transformer model which has 24 self-attention encoder layers and 8 decoder layers.

Table~\ref{tab:systems} shows the offline performance of all the investigated S2S models in this work. The big LSTM model achieved the best WER performance while the transformer performs worse. However, the transformer is very efficient to supplement the LSTM models in the combination. The ensemble of 3 models (labeled as \textit{E3}) results in a single system that achieved a 5.0\% WER on the SWB test set, which is on par with the state-of-the-art performance on this benchmark.
\vspace{-0.0cm}

\vspace{-0.0cm}
\subsection{Latency with Shared Prefix}
\label{ssec:result_latency_shared_prefix}
We use an audio chunk size of 300ms to perform incremental inference with the systems in Table~\ref{tab:systems}. All inferences were performed on a single Nvidia Titan RTX GPU. Table~\ref{tab:share_prefix} shows the WERs for SWB, \textit{computation latency} and \textit{confidence latency} (see Section~\ref{sec:latency_measure}) for different beam sizes when only using the \textit{share prefix} strategy for stability detection.

As can be seen, the confidence latency is much larger than the computation latency in all the experiments and shown to be a more critical factor for final latency improvement. The systems involving multiple S2S models require more computational power, however, they obtain better confidence latency and accuracy due to the reduction of model uncertainty.

When using a high beam size (e.g., 8), all the experimental systems can achieve their offline accuracy. This result reveals interesting observations for making online S2S ASR systems. First, as this condition is reliable among different S2S architectures, it shows that all S2S ASR models may share the same characteristic in which they tend not to use further context for the inference of a given prefix at a particular time. This observation is consistent with the finding in \cite{nguyen2020high} for the LSTM-based S2S model. Secondly, it proves that the use of bidirectional encoders in online conditions is possible and even results in the same optimal accuracy as in offline inference. Lastly, it reveals a unified approach to build online ASR for different S2S architectures. As an attractive advantage, this approach does not require model modifications. 

The best system using the shared prefix condition achieved a WER of 5.0\% and suffered an average delay of 1.79 seconds which is slightly slower than the one with lowest latency.

\begin{table}[t]
	\caption{Computation and confidence latency when using shared prefix condition.}
	\label{tab:share_prefix}
	\setlength{\tabcolsep}{5pt}
	\centering
	\begin{tabular}{|c|c|c|c|c|c|}
		\hline
		Model & Beam Size & Comp. & Conf. & SWB \\
		\hline \hline
		    S1 & 8 & 0.10 & 1.50 & 5.8 \\
		    S2 & 8 & 0.13 & 1.55 & 5.6 \\
    		T1 & 8 & 0.19 & 1.50 & 5.8 \\
    		T1 & 6 & 0.16 & 1.35 & 5.9 \\
    		T1 & 4 & 0.12 & 0.70 & 6.6 \\
    		\hline
    		E1 & 8 & 0.18 & 1.55 & 5.3 \\
    		E2 & 8 & 0.29 & 1.50 & 5.0 \\
    		E2 & 6 & 0.25 & 1.30 & 5.1 \\
    		E2 & 4 & 0.20 & 0.80 & 5.7 \\
		\hline
	\end{tabular}
	\vspace{-0.2cm}
\end{table}

\vspace{-0.0cm}
\subsection{Trade-off for Better Latency}
\label{ssec:result_latency_trade_off}
\begin{figure}[t]
	\centering
	\vspace{-0.0cm}
	\includegraphics[width=0.98\linewidth]{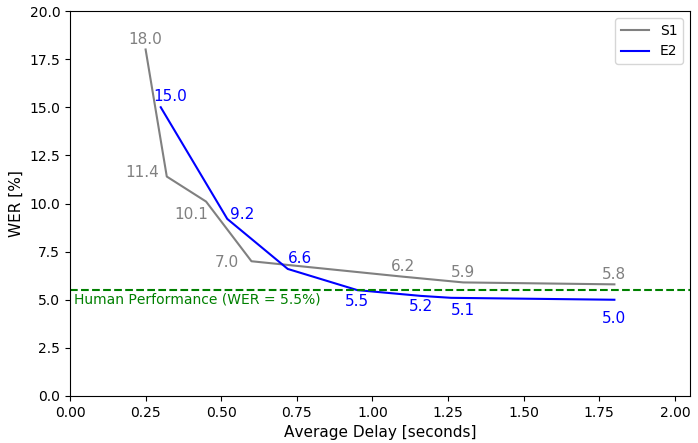}
    \vspace{-0.0cm}
	\caption{Trade-off between latency and accuracy. Beam size of 8 is used for both systems.}
	\label{fig:tradeoff}
	\vspace{-0.2cm}
\end{figure}

To further improve the latency, we use both the stability detection strategies from Section \ref{ssec:stability_dectect}. We do the combination via a logical OR which means the stability is detected as soon as one of the conditions applies. At the end, we can trade-off latency against accuracy as the function of the term $\Delta$ -- the delay time needed to finalize the endpoint of a prefix. Figure~\ref{fig:tradeoff} presents the trade-off curves for two systems, \textit{S1} and \textit{E2}. In both systems, the model \textit{S1} is used for detecting the \textit{best-ranked prefix} condition.

As can be seen, both systems can achieve much better latency (of only 1.30 seconds) with only a slight increase in WER (e.g., 0.1\% absolute). The ensemble system \textit{E2} achieves a latency of 0.85 seconds while yielding the same accuracy as \textit{S1}. Human performance (5.5\%) can be reached with an average delay of only 1 second. Note that, the WER for human performance was extracted as the average of the two studies \cite{xiong2016achieving} and \cite{saon2017english}.

\vspace{-0.0cm}
\subsection{Compared to Other Works}
Table \ref{tab:comparision} presents the WER performance from recent studies for online and offline conversational speech recognition systems. Human WER performance was obtained in 2016 with a combination of different HMM hybrid ASR systems. While until 2019 and 2020, new records on offline conversational speech was set with end-to-end sequence-to-sequence ASR systems. We found only a few attempts \cite{audhkhasi2019forget,kurata2020knowledge} that make streaming ASR for this benchmark. In these studies, the accuracy between offline and streaming conditions was shown to be in clear margins. In a different manner, we show the offline accuracy can be possibly reached with our proposed low-latency S2S system. Our best achieved online WER is slightly behind the state-of-the-art offline performance on the Switchboard benchmark.

\begin{table}[h]
	\caption{Results from other works on SWB test set.}
	\label{tab:comparision}
	\setlength{\tabcolsep}{4pt}
	\centering
	\begin{tabular}{|l|c|c|c|c|}
		\hline
		Model & Train. Data & Condition & WER \\
		\hline \hline
		Hybrid \cite{saon2017english} (2017) & SWB+Fisher & \textit{Offline} & 5.5 \\
		Hybrid \cite{xiong2018microsoft} (2018) & SWB+Fisher & \textit{Offline} & 5.1 \\
		S2S \cite{nguyen2020improving} (2019) & SWB+Fisher & \textit{Offline} & 5.2 \\
		S2S \cite{wang2020investigation} (2020) & SWB+Fisher & \textit{Offline} & 4.9 \\
		S2S \cite{tuske2020single} (2020) & SWB+Fisher & \textit{Offline} & 4.8 \\		
		CTC \cite{audhkhasi2019forget} (2019) & SWB & \textit{Streaming} & 9.1 \\
		Transducer \cite{kurata2020knowledge} (2020) & SWB & \textit{Offline} & 12.8 \\
		Transducer \cite{kurata2020knowledge} (2020) & SWB & \textit{Streaming} & 17.0 \\
		\hline
		\textit{Ours} & SWB+Fisher & \textit{Low-latency} & 5.0 \\
		\hline
	\end{tabular}
\end{table}
\vspace{-0.0cm}

\vspace{-0.25cm}
\section{Conclusion}
\label{sec:conclusion}
We have shown a unified approach to construct online and low-latency ASR systems for different S2S architectures. The proposed online system employing three S2S models works either in an accuracy-optimized fashion that achieves state-of-the-art performance on telephone conversation speech or in a very low-latency manner while still producing the same or better accuracy as the reported human performance.

\section{Acknowledgement}
The project ELITR leading to this publication has received funding from the European Unions Horizon 2020 Research and Innovation Programme under grant agreement No 825460.

\bibliographystyle{IEEEtran}

\bibliography{mybib}

\end{document}